\documentclass[conference]{IEEEtran}
\IEEEoverridecommandlockouts
\usepackage[american]{babel}

\usepackage[numbers, compress]{natbib} 

\usepackage{subcaption}
\usepackage[]{algorithm2e}
\SetKwIF{If}{ElseIf}{Else}{if}{}{else if}{else}{end if}

\RestyleAlgo{ruled}

\usepackage{mathtools}
\usepackage{nccmath}
\usepackage{multirow}
\usepackage{arydshln}
\usepackage{amssymb}
\usepackage{url}
\usepackage{adjustbox}
\usepackage{wrapfig}
\usepackage{float}
\usepackage{hyperref}

\usepackage{wrapfig,lipsum,booktabs}

\newcommand{\vecX}{\mathbf{x}}
\newcommand{\vecW}{\mathbf{w}}
\newcommand{\vecP}{\mathbf{p}}
\newcommand{\vecH}{\mathbf{h}}
\newcommand{\vecF}{\mathbf{f}}

\newcommand{\expect}{\mathbb{E}}

\begin{document}
\title{WWAggr: A Window Wasserstein-based Aggregation for Ensemble Change Point Detection}

\author{\IEEEauthorblockN{1\textsuperscript{st} Alexander Stepikin}
\IEEEauthorblockA{\textit{Applied AI Center, Skoltech} \\
Moscow, Russia \\
alexander.stepikin@skoltech.ru}
\and
\IEEEauthorblockN{2\textsuperscript{nd} Evgenia Romanenkova}
\IEEEauthorblockA{\textit{Applied AI Center, Skoltech} \\
Moscow, Russia \\
shulgina@phystech.edu}
\and
\IEEEauthorblockN{3\textsuperscript{rd} Alexey Zaytsev}
\IEEEauthorblockA{\textit{Applied AI Center, Skoltech}\\
 Moscow, Russia\\
\textit{BIMSA}, \\
Beijing, China}}


    \maketitle

\begin{abstract}
Change Point Detection (CPD) aims to identify moments of abrupt distribution shifts in data streams. 
Real-world high-dimensional CPD remains challenging due to data pattern complexity and violation of common assumptions. 
Resorting to standalone deep neural networks, the current state-of-the-art detectors have yet to achieve perfect quality. 
Concurrently, ensembling provides more robust solutions, boosting the performance. 
In this paper, we investigate ensembles of deep change point detectors and realize that standard prediction aggregation techniques, e.g., averaging, are suboptimal and fail to account for problem peculiarities.
Alternatively, we introduce WWAggr --- a novel task-specific method of ensemble aggregation based on the Wasserstein distance. 
Our procedure is versatile, working effectively with various ensembles of deep CPD models. 
Moreover, unlike existing solutions, we practically lift a long-standing problem of the decision threshold selection for CPD.
\end{abstract}


\section{Introduction}
\label{sec:introduction}
\emph{Change Point Detection} (CPD) addresses the challenge of precise identification of the moments when some statistical data distribution properties undergo alterations.
Such a problem emerges in various real-world scenarios: manufacturing process monitoring~\cite{lattari2022deep, bao2024self}, server logs~\cite{tran2019automated}, 
financial data analysis~\cite{laptev2015s5, qin2025pca}, or video surveillance~\cite{romanenkova2022indid}.
In these situations, an unexpected change often indicates an emergency that requires an immediate response.
Thus, developing accurate CPD methods is essential to ensure the safe deployment of automatic systems in applications.

CPD possesses a rich theoretical foundation with optimal procedures proposed for both online and offline detection~\cite{pollak2009optimality,truong2020selective}.
However, these methods rely on strong assumptions about data distributions, input dimensionality, or sparsity of the change signal, which limits their applicability to real-world data~\cite{xie2021sequential}.
In response, various deep CPD models have been proposed~\cite{lattari2022deep, romanenkova2022indid, chang2018kernel, deldari2021time,bazarova2024norm,ryzhikov2023latent,xu2025change}, targeting practical demands and offering data-driven solutions with fewer prior assumptions.

While state-of-the-art (SOTA) deep CPD models demonstrate near-perfect performance for simple, low-dimensional time series, they often struggle when dealing with complex, multivariate data, like video streams~\cite{romanenkova2022indid, ryzhikov2023latent}.
This observation suggests that individual models may lack the expressive power required to capture the diverse range of change points (CPs) occurring in the real world.
A potential solution to this problem is ensembling, which has proven to boost weak standalone predictors~\cite{lee2015m} and enhance model expressiveness~\cite{wang2023diversity}.
Moreover, the inconsistency of ensemble base learners with each other provides natural uncertainty estimation~\cite{wang2023diversity,lakshminarayanan2017simple, kotelevskii2025multidimensional} and may serve as an additional source of information for identifying CPs.
Despite current advancements, recent CPD ensembling~\cite{qin2025pca,artemov2015ensembles, katser2021unsupervised} still relies on simple base detectors unsuitable for high-dimensional data, ignores the task's specifics, and disregards the issue of optimal alarm threshold selection.

\begin{figure*}[ht]
\centering
\includegraphics[width=0.7\textwidth]{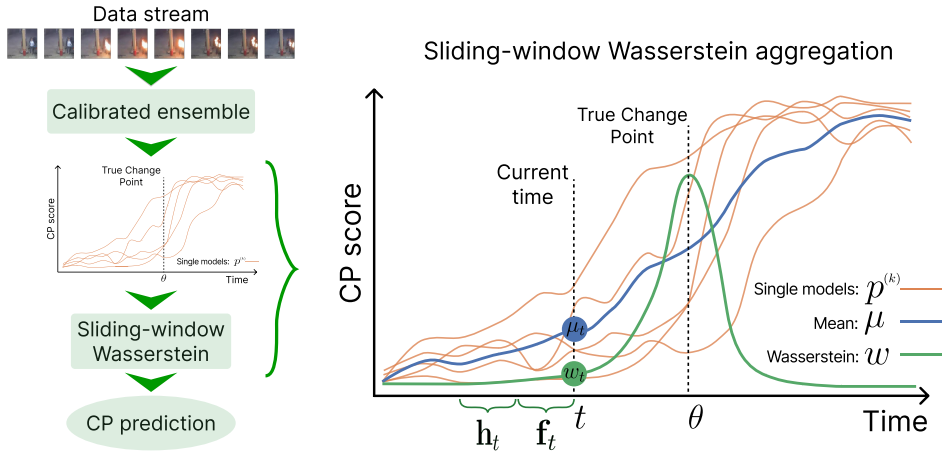}
\caption{
Teaser of our new aggregation procedure for effective high-dimensional change point detection.
First, an ensemble of deep change point (CP) detectors predicts well-calibrated CP scores for each moment.
Second, these scores are aggregated via WWAggr --- our sliding-window Wasserstein procedure.  
Applied to the calibrated scores, WWAggr better reflects changes in ensemble predictions, yielding near-optimal performance for the pre-defined threshold of $0.5$.
}
\label{fig:pipeline}
\end{figure*}

To tackle these challenges, we introduce a sliding-Window Wasserstein-based Aggregation (WWAggr) technique for ensemble CP predictions.
The general workflow (Figure~\ref{fig:pipeline}) includes training an ensemble of deep models to predict CP scores and aggregating these predictions with our CPD-specific technique.
This algorithm is model-agnostic and robust to alarm threshold selection when the ensemble base learners are well-calibrated~\cite{rahimi2020post, wang2023calibration}.
For WWAggr, any reasonable threshold gives close to optimal performance, so one can select this value in advance without a subsequent quality drop, increasing the procedure's utility in online applications.

In summary, our main contributions are the following.
\begin{enumerate}
    \item We propose WWAggr --- a novel CPD-specific ensemble aggregation technique that leverages the Wasserstein distance.
    Our procedure is model-agnostic: it successfully works with ensembles of various deep CP detectors.
    \item To enhance the robustness against the alarm threshold selection, we incorporate model calibration. 
    This enables a crucial practical advantage: the model operates effectively for any reasonable threshold near $0.5$, facilitating online inference.
    \item Experiments with different deep CP detectors (supervised/unsupervised) on real-world datasets, including video surveillance, show that WWAggr-aggregated ensembles outperform standalone models and conventionally aggregated ensembles (Table~\ref{tab:ranks}), establishing new SOTA, especially for complex high-dimensional setting.

\end{enumerate}


\begin{table}[!ht]
\caption{
Mean performance ranks for the considered aggregation procedures.
The ranks are averaged across all the models and datasets under study.
For standalone models with no aggregation of predictions, the results are marked as ``s.m.''
The best value is highlighted in \textbf{bold}.
}
\label{tab:ranks}
\centering
\normalsize{
\begin{tabular}{l|c}
\hline
 Aggregation & Mean $F_1$ rank \\ \hline
 None (s.m.) & 4.8 \\\hdashline
 Mean & 3.1 \\
 Min & 5.1 \\
 Max & 3.3 \\
 Median & 3.0 \\
 WWAggr (ours) & \textbf{1.2} \\
\hline
\end{tabular}
}
\end{table}

The entire pipeline implementation, including ensemble training, calibration, and WWAggr is available online at \url{https://github.com/stalex2902/wwagr-ensemble-cpd}.
\section{Related work}
\label{sec:literature_review}
\paragraph{Change Point Detection}
CPD is well-studied in statistics and machine learning~\citep{xie2021sequential,van2020evaluation}.
Originating from control charts~\citep{page1954continuous}, there exist classic CPD procedures that are proven optimal for low-dimensional time series~\citep{pollak2009optimality, xie2021sequential}.
However, they have severe practical limitations.
For instance, CUSUM or Shiryaev-Roberts statistics require prior knowledge of data distribution before and after a change point~\citep{shiryaev2017stochastic}.
Nevertheless, these methods continue to evolve, accommodating a broader range of data models~\citep{xie2021sequential}.
Another approach, distance-based CPD, detects CPs by comparing the probability distributions of consecutive data segments.
The typical test statistics here are based on probabilistic measures, e.g., the Maximum Mean Discrepancy~\citep{gretton2012kernel,li2015m} or the Wasserstein distance~\citep{ramdas2017wasserstein,faber2021watch}. 
While effective for simple series, these methods face challenges with multivariate data: the ``curse of dimensionality'' and resource-intensive estimation of distances~\citep{qin2025pca,enikeeva2019high,chen2022high, zhang2023projection}.

Deep CPD~\citep{xu2025change} overcomes these problems and provides the current SOTA for high-dimensional data streams. 
Typically, these methods embed the data into low-dimensional representation space instead of operating on raw time series~\citep{deldari2021time,bazarova2024norm,wang2018high}. 
The power of modern network architectures allows us to learn embeddings that accurately reflect the statistical features of the original series.  
Informative representations can be obtained in both supervised~\citep{lattari2022deep, romanenkova2022indid} and unsupervised~\citep{deldari2021time, puchkin2023contrastive} regimes.
Moreover, current research suggests that incorporating principled CPD practices into neural network training enhances detection effectiveness.
For example, scholars parametrize classic CPD approaches with neural networks~\citep{chang2018kernel, hushchyn2020online} or use specific loss functions~\citep{romanenkova2022indid,lee2022neural}. 
However, deep CP detectors constitute standalone networks with limited generalization~\cite{ganaie2022ensemble} that are still far from optimal performance for real-world multivariate data.

\paragraph{Ensembling}
In deep learning, ensembling involves training several neural networks with varied weight initializations, training data subsets, or hyperparameters~\citep{lee2015m,ganaie2022ensemble}. 
These basic approaches show the potential of ensembles to leverage diverse patterns and improve results by aggregating information from multiple learners.

Ensembling has also been explored within the CPD framework. 
Combining traditional CPD methods~\cite{qin2025pca,artemov2015ensembles,katser2021unsupervised}, recent studies consistently demonstrate that ensembles outperform individual baselines. 
Ensembling strategies include leveraging the properties of diverse detectors~\cite{katser2021unsupervised}, optimizing CPD-specific criteria for ensemble construction~\cite{artemov2015ensembles}, or applying univariate CPD methods to each dimension of data projections~\cite{qin2025pca}. 
Notably, the authors of~\cite{qin2025pca} emphasize the importance of aggregating individual model outputs to refine the final decision, though their own approach utilizes simple median aggregation. 
To summarize, current ensemble methodologies in CPD remain focused on traditional algorithms, thereby overlooking recent advancements in neural CPD and more sophisticated, task-specific ensemble aggregation techniques. 

\paragraph{Finalizing ensemble predictions}
A crucial step in applying ensemble methods is selecting an aggregation strategy for the final decision.
Current approaches in CPD~\citep{artemov2015ensembles,katser2021unsupervised} use simple aggregations like averaging predictions or taking maximum/minimum values. 
In contrast, within the related field of anomaly detection, research shows that disagreements in ensemble predictions can indicate an anomaly score~\citep{lakshminarayanan2017simple}, highlighting the need for task-specific aggregation in CPD.

Another key aspect of finalizing deep ensemble predictions is model calibration.
Most researchers find the calibration enhances empirical performance~\citep{kotelevskii2025multidimensional, stickland2020diverse, wen2020combining}.
The logic behind this step is that diverse base learners may produce outputs with different distributions, thereby requiring some transformation before final decision-making. Additionally,~\cite{krishnan2020improving} shows that proper model calibration increases the discrepancy between normal and abnormal data in anomaly detection.
We hypothesize that it can be beneficial for CPD in a similar manner and, thus, suggest using post-hoc calibration~\citep{rahimi2020post}, which applies deterministic transformations to model outputs. 
The key methods here include Platt Scaling~\citep{platt1999probabilistic}, Temperature Scaling~\citep{guo2017calibration}, and beta calibration~\citep{kull2017beyond}. 

\paragraph{Resume} 
Real-world tasks like video monitoring demand effective high-dimensional CPD. 
Classic methods fall short in this scenario, necessitating the use of neural detectors. 
Meanwhile, existing research typically focuses on individual deep models, which often yield suboptimal performance, or explores ensembles of classic CPD algorithms, ill-suited for complex data.
This paper investigates ensembles of SOTA deep change point detectors. 
We explore how proper ensemble output aggregation tailored to CPD and model calibration enhances detection quality and usability for practical tasks.
\section{Problem statement}
\label{sec:problem_statement}
Let $X_{1:T} = \{\vecX_{t}\}_{t = 1}^{T}$, $\vecX_{t} \in \mathbb{R}^D$, denote a multivariate time series. 
Suppose there exists a moment $\theta \in[1, T]$ such that for $t < \theta$, the observations $\vecX_{t}$ follow a ``normal-data'' distribution $f_{0}$, and for $t \geq \theta$ the distribution of $\vecX_{t}$ switches to $f_{1}$ --- an ``abnormal'' one.
In this case, $\theta$ is called a \emph{change point} (CP).
CPD aims to identify the appearance of $\theta$ quickly and accurately, producing its estimate $\tau$.
Notably, we operate under a common ``at-most-one-change'' assumption~\citep{shiryaev2017stochastic}, which can be extended to the multiple CPD scenario.

As with any distance-based procedure, our pipeline is inspired by two-sample hypothesis testing. 
In this setting, one determines whether two consecutive data subsets are generated from a single probability distribution~\citep{chang2018kernel}:
\begin{gather}
    \label{eq:hypothesis_2sample}
    \mathcal{H}_{0}^{(t)}\colon f_{X_{t-\omega:t}} = f_{X_{t:t+\omega}}; \quad
    \mathcal{H}_{1}^{(t)}\colon f_{X_{t-\omega:t}} \neq f_{X_{t:t+\omega}}.
\end{gather}
\noindent Here, $X_{t-\omega:t} = \{\vecX_{t^{\prime}}\}_{t^{\prime}=t-\omega}^{t-1}$ and $X_{t:t+\omega} = \{\vecX_{t^{\prime}}\}_{t^{\prime}=t}^{t+\omega-1}$ are two consecutive sub-series of size $\omega$ for some step $t$, and $f_X$ denotes the probability density function for a sample $X$.
Eventually, $\tau$ is the first moment when $\mathcal{H}_{0}^{(t)}$ is rejected at some confidence level.

Targeting high-dimensional data, our pipeline (Section~\ref{sec:methods}) uses deep change point detectors that estimate $f_{0,1}$ by learning data representations.
Further, we operate in the space of the predicted CP scores, detecting distributional shifts there.
\section{Methods}
\label{sec:methods}
This section presents the proposed technique for the aggregation of ensemble-based CP predictions.
We begin with general sequence-to-sequence (seq2seq) CPD ensembles and then proceed to the details of further steps in our pipeline.

\subsection{Deep CPD ensembles}
\label{subsec:cpd_ensembles}
We explore ensembles of deep CP detectors operating in a seq2seq mode.
In particular, each base learner is a neural network $g_{\vecW}$ with parameters $\vecW$ that takes the series $X_{1:T}$ as an input and outputs the sequence of the predicted CP scores: $g_{\vecW}\left(X_{1:T}\right) = p_{1:T} = \{ p_t \}_{t = 1}^{T}$.
If $K$ models are trained independently, we obtain an ensemble $\left\{g_{\vecW_{k}} \right\}_{k = 1}^{K}$.

To enforce model diversity and, thereby,  boost the overall ensemble expressiveness, several techniques during ensemble training are typically used~\citep{ganaie2022ensemble}:
(1) setting different initializations for the models' weights and the stochastic gradient descent; 
(2) fitting the models on bootstrapped subsamples of the initial training dataset;
(3) using specific noisy optimization procedures, like SGLD~\citep{welling2011bayesian}.
Our sensitivity studies (see \ref{subsec:sensitivity_studies}) have shown that the most straightforward option (1) provides the superior solution with a sufficient variety of base learners.
Thus, this option is used in the main experiments.

\subsection{Ensemble prediction aggregation}
\label{subsec:ensemble_prediction_agregation}
Once the ensemble is trained and the predictions $P_{1:T} = \left\{p_{1:T}^{(k)}\right\}_{k=1}^{K}$ are obtained, the simplest way to derive the final decision is to average the outputs~\citep{qin2025pca,ganaie2022ensemble}:
$\mu_{1:T} = \frac{1}{K}\sum_{k=1}^{K}{p_{1:T}^{(k)}}$.
Alternatively, \cite{katser2021unsupervised} addresses other ``naive'' aggregations (e.g., min, max, sum, or quantile values).
However, none of them are inherently designed for CPD, nor do they account for the problem's peculiarities.
We expect that, similarly to the anomaly detection area~\citep{lakshminarayanan2017simple,krishnan2020improving}, the heterogeneity of base learners' predictions could serve as another vital, yet often underexplored, source of information about the CP location. This hypothesis inspires \emph{WWAggr}, our task-specific aggregation procedure.

\begin{algorithm}[ht]
\SetKwInOut{Input}{Input}\SetKwInOut{Output}{Output}
\Input{
    $\{g_{\vecW_{k}}\}_{k=1}^{K}$ --- an ensemble of trained and calibrated CPD models; \\
    $X_{1:T}$ --- a multivariate time series; \\ 
    $d(\cdot, \cdot)$ --- a probabilistic distance function; \\
    $\omega$ --- a window size; 
    $h$ --- an alarm threshold.
}
\Output{$\tau$ --- change point prediction.}
\tcc{get ensemble predictions}
Compute $p_{1:T}^{(k)} = g_{\vecW_{k}}\left(X_{1:T}\right)$ for $k=1,\ldots, K$.\\ 
\tcc{aggregate ensemble predictions}
Set $w_{1} = \ldots = w_{2\omega} = 0$.\\
\For{$t = 2\omega + 1$ $\operatorname{to}$ $T$}{
    1) Obtain ``future'' $F_{t} = \vecP_{t-\omega:t}$ of size $\omega\times K$. \\
    2)  Obtain ``history'' $H_{t} = \vecP_{t-2\omega:t-\omega}$ of size $\omega\times K$. \\
    3) Flatten $F_{t}$ and $H_{t}$ into the vectors $\vecF_{t}$ and $\vecH_{t}$ of size~$1\times \omega K$. \\
    4) Compute $w_{t} = d(\vecF_{t}, \vecH_{t})$. \\
}
\tcc{get the final CP estimate}
\uIf{$\forall t\in\overline{1, T}\colon w_{t} < h$}{$\tau = T$}
\uElse {$\tau = \min\{t\colon w_{t} \geq h\}$}
\Return $\tau$
\caption{WWAggr for ensemble CPD}
\label{alg:distance_procedure}
\end{algorithm}

\paragraph{WWAggr procedure}
More precisely, we propose an online sliding window procedure inspired by two-sample statistical tests common in CPD.
WWAggr iterates through the series of ensemble predictions $P_{1:T}$ with two windows of size $\omega$: ``history'' $H_{t}$ and ``future'' $F_{t}$. 
By computing the probabilistic distance $d(\cdot, \cdot)$ between them, we detect the distribution shift in the ensemble's CP scores $P_{t}$ and raise the alarm when this distance exceeds a predefined threshold $h$.
While similar in concept to~\citep{faber2021watch}, WWAggr operates on the ensemble's output predictions rather than the raw time series data. 
Importantly, compared to standard ``mean'' aggregation, WWAggr leverages all the information provided by an ensemble, including model uncertainty.
Please find further details in Algorithm~\ref{alg:distance_procedure}.

\paragraph{Choice of the probabilistic distance}
Although there are several theoretically sound options to measure distribution similarity, we settle on the \emph{Wasserstein distance}~\citep{ramdas2017wasserstein} for the reasons discussed below.

The $p$-Wasserstein distance between two $d$-dimensional probability distributions, $\mathbb{P}$ and $\mathbb{Q}$, is defined as:
\begin{equation}
\label{eq:wasserstein}
    W_{p}(\mathbb{P}, \mathbb{Q}) = \left(\inf_{\pi\in\Gamma\left(\mathbb{P}, \mathbb{Q}\right)}{\int_{\mathbb{R}^{d}\times\mathbb{R}^{d}}{\|\mathbf{x} - \mathbf{y}\|^{p} d\pi}}\right)^{\frac{1}{p}},
\end{equation}
where $\Gamma\left(\mathbb{P}, \mathbb{Q}\right)$ is the set of joint probability measures on $\mathbb{R}^{d}\times\mathbb{R}^{d}$ with marginals $\mathbb{P}$ and $\mathbb{Q}$, and $p\in\mathbb{N}$ is the order parameter.
While being difficult to compute in the general case, this distance is much simpler when $d = p = 1$: 
\begin{equation}
\label{eq:wasserstein_1d_invert}
    W_{1}(\mathbb{P}, \mathbb{Q}) = \int_{\mathbb{R}}{\left|F_{\mathbb{P}}(x) - F_{\mathbb{Q}}(x)\right|dx},
\end{equation}
with $F_{\mathbb{P}}$ and $F_{\mathbb{Q}}$ being the corresponding cumulative density functions (CDFs).
Substituting true CDFs with the empirical ones, we can estimate the 1-Wasserstein distance between the given samples $X^{(n)}\sim \mathbb{P}$ and $Y^{(n)}\sim \mathbb{Q}$ of size $n$ in the following way:
\begin{equation}
\label{eq:wasserstein_1d_estimator}
    \hat{W}_{1}\left(X^{(n)}, Y^{(n)}\right) = \frac{1}{n}\sum_{i=1}^{n}{\left|X^{(n)}_{(i)} - Y^{(n)}_{(i)}\right|}, 
\end{equation}
where $X^{(n)}_{(i)}$ and $Y^{(n)}_{(i)}$ denote the $i$-th order statistics of the corresponding samples.
The work~\cite{klein2017convergence} proves the strong consistency of this estimator for a broad family of continuous distributions, thereby justifying its practical usage.

While the 1-Wasserstein distance could also be estimated for $d > 1$, the results from~\cite{weed2019sharp} established that the convergence speed of this estimator to the true distance is $O\left(n^{-1/d}\right)$.
Consequently, with the fixed ensemble of $K$ models, a faster convergence can be achieved if we use one-dimensional ($d = 1$) samples of size $n = \omega K$ rather than estimate the distance between two $K$-dimensional samples, each of size $\omega$.
Moreover, employing a moderate window size $\omega$ is generally preferred, as it results in a smaller inherent detection delay.
This observation motivates us to introduce the flattening step within Algorithm~\ref{alg:distance_procedure}.
The decision is also justified by the assumption that after calibration (discussed below), CP scores predicted by distinct models come from the same distribution.

The \emph{Maximum Mean Discrepancy} (MMD) is another probabilistic distance widely used for CPD~\citep{gretton2012kernel, li2015m}.
As the Wasserstein metric can be interpreted as a specific case of the MMD~\citep{ramdas2017wasserstein},
they exhibit similar performance regarding the quality metrics if the alarm threshold is selected properly (see \ref{subsubsec:choice_of_probabilistic_distance}).
However, the 1-Wasserstein distance possesses a valuable property of being bounded to $[0, 1]$ (see below).
This feature, combined with the proper model calibration, alleviates the problem of the alarm threshold selection (see Section~\ref{subsec:results_calibration}).

\paragraph{Optimal alarm threshold selection}
We employ seq2seq CPD models that predict a CP score $p_{t}$ for each time step $t$. 
A CP is reported when $p_{t}$ exceeds a predefined alarm threshold $h$ for the first time. 
The same applies to the ensemble models with naive aggregations and our CPD-specific procedure discussed above.
Since CP detector performance heavily depends on the choice of hyperparameters that control alarm signal~\citep{van2020evaluation}, it is essential to thoroughly estimate the optimal threshold value on a hold-out set before model inference.

In our Algorithm~\ref{alg:distance_procedure}, the samples $X^{(n)}$ and $Y^{(n)}$ are the flattened history and future windows $\vecH_{t}$ and $\vecF_{t}$.
They consist of the predicted CP scores $P_{1:T}$, bounded to $[0, 1]$ due to the nature of the considered base models (see Appendix~\ref{sec:impl_details}).
Thus, for any $i$, $0 \leq \left|X_{(i)} - Y_{(i)}\right| \leq 1$, and, consequently, $0\leq w_{t} = d(\vecF_{t}, \vecH_{t}) = \hat{W}_{1}\left(X^{(n)}, Y^{(n)}\right) \leq 1$.
Therefore, for WWAggr, meaningful threshold values belong to $[0, 1]$, making our aggregation convenient in practice.
Simultaneously, there are no such guarantees for MMD.

Practically, reducing sensitivity to hyperparameters, particularly $h$, is highly desirable. 
The ideal CPD procedure should work effectively enough for a fixed alarm threshold, e.g., $0.5$.
Our findings indicate that for the WWAggr method, this feature can be achieved through proper model calibration.

\paragraph{CPD model calibration}
Correct model calibration was found to be essential in anomaly detection, as it allows better distinguishing between normal and abnormal data distributions~\citep{krishnan2020improving}.
We expect similar behavior from the CPD and hypothesize that this effect could benefit WWAggr.
Thus, we research the calibration procedures for the CPD models.

As shown by~\cite{lattari2022deep, romanenkova2022indid, hushchyn2020online}, CPD can be interpreted as a seq2seq binary classification task.
It allows us to follow~\cite {kull2017beyond} and use model calibration notation designed for binary classifiers. 
Recall that we work with the baseline models of the form $p_{t} = g_{\vecW}\left(X_{1:t}\right)$.
The model $g_{\vecW}$ is perfectly calibrated if
$p_{t} = \expect\left[l_{t}|g_{\vecW}\left(X^{1:t}\right) = p_{t}\right]$,
where $p_{t}$ is a predicted CP score, serving as the positive class probability, and $l_{t}$ is the true binary label ($l_{t} = 0$ for $t < \theta$, $l_{t} = 1$ for $t\geq \theta$).
This equation means that if a well-calibrated model predicts probability $p$, the actual fraction of positives corresponding to this model output should equal $p$.

To present a unified framework, independent of base model training procedures, we resort to \emph{post-hoc calibration} methods, which assume that the model of interest has already been trained and its parameters are frozen.
Therefore, better calibration is achieved by post-processing the model's outputs.
Let $s$ denote the model's logit for a particular input. 
The predicted probability $p$ is then commonly computed as the sigmoid function: $p(s) = \sigma(s) = (1 + \exp(-s))^{-1}$.
We settle on the beta calibration method~\citep{kull2017beyond}, where the Beta distribution inspires the form of the logit transformation:
\begin{equation}
\label{eq:beta_calibration}
p_{\mathrm{beta}}\left(s; a, b, c\right) = \left(1 + \frac{\left(1 - s\right)^{b}}{s^{a}}e^{-c} \right)^{-1}
\end{equation}
for parameters $a, b\in\mathbb{R}$, and $c\geq 0$ selected by optimizing the cross-entropy on the hold-out set.
Despite its simplicity, this method presents a firm baseline in post-hoc calibration~\citep{wang2023calibration}.
Our study (Subsection~\ref{subsubsec:sens_calibration_method}) illustrates that it works well for CPD, outperforming Temperature Scaling~\citep{guo2017calibration} --- another popular post-hoc calibration approach.
\section{Experiments}
\label{sec:experiments}
Our final pipeline follows Algorithm~\ref{alg:distance_procedure}, with $d(\cdot,\cdot)$ being the 1-Wasserstein distance and base learners $g_{\vecW_{k}}$ being the beta-calibrated deep CPD models.
In this section, we evaluate our approach in real-data scenarios with various underlying base detectors for both supervised and unsupervised scenarios.

\begin{table*}[ht]
\caption{
Detection $F_1$-scores for the ensembles of supervised and unsupervised CPD models.
The results are averaged by three runs and are given in the format $mean \pm std$.
For standalone models, no prediction aggregation is used and these results are marked as ``s.m.''
The best values are highlighted in \textbf{bold}; the second best are \underline{underlined}.
}
\label{tab:results_ens}
\centering
\begin{tabular}{l|c|cc|cc}
\hline
\multirow{2}{*}{Dataset} & \multirow{2}{*}{Aggregation} & \multicolumn{2}{c|}{Supervised models} & \multicolumn{2}{c}{Unsupervised models} \\ \cline{3-6}
 & & BCE & InDiD & TS-CP$^{2}$ & SN-TS2Vec \\ \hline
\multirow{6}{*}{Yahoo} & None (s.m.) & 0.895 \scriptsize{$\pm$ 0.038} & 0.871 \scriptsize{$\pm$ 0.023} & 0.855 \scriptsize{$\pm$ 0.034} & \underline{0.774} \scriptsize{$\pm$ 0.033} \\ \cdashline{2-6}
 & Mean & 0.892 \scriptsize{$\pm$ 0.029} & 0.873 \scriptsize{$\pm$ 0.025} & 0.872 \scriptsize{$\pm$ 0.011} & 0.765 \scriptsize{$\pm$ 0.022} \\
 & Min & 0.872 \scriptsize{$\pm$ 0.043} & 0.867 \scriptsize{$\pm$ 0.036} & 0.851 \scriptsize{$\pm$ 0.018} & 0.753 \scriptsize{$\pm$ 0.008} \\
 & Max & \underline{0.901} \scriptsize{$\pm$ 0.028} & \textbf{0.888} \scriptsize{$\pm$ 0.023} & 0.865 \scriptsize{$\pm$ 0.005} & 0.738 \scriptsize{$\pm$ 0.025} \\
 & Median & \textbf{0.908} \scriptsize{$\pm$ 0.019} & 0.878 \scriptsize{$\pm$ 0.023} & \underline{0.873} \scriptsize{$\pm$ 0.004} & 0.765 \scriptsize{$\pm$ 0.047} \\
 & WWAggr & \underline{0.901} \scriptsize{$\pm$ 0.027} & \underline{0.882} \scriptsize{$\pm$ 0.026} & \textbf{0.891} \scriptsize{$\pm$ 0.021} & \textbf{0.785} \scriptsize{$\pm$ 0.016} \\
\hline
\multirow{6}{*}{Explosions} & None (s.m.) & 0.695 \scriptsize{$\pm$ 0.058} & 0.560 \scriptsize{$\pm$ 0.070} & 0.498 \scriptsize{$\pm$ 0.080} & 0.535 \scriptsize{$\pm$ 0.053} \\ \cdashline{2-6}
 & Mean & 0.701 \scriptsize{$\pm$ 0.052} & 0.588 \scriptsize{$\pm$ 0.019} & \underline{0.587} \scriptsize{$\pm$ 0.044} & 0.563 \scriptsize{$\pm$ 0.026} \\
 & Min & 0.679 \scriptsize{$\pm$ 0.047} & 0.568 \scriptsize{$\pm$ 0.087} & 0.574 \scriptsize{$\pm$ 0.019} & \textbf{0.564} \scriptsize{$\pm$ 0.061} \\
 & Max & \underline{0.735} \scriptsize{$\pm$ 0.050} & \underline{0.593} \scriptsize{$\pm$ 0.027} & 0.565 \scriptsize{$\pm$ 0.047} & 0.563 \scriptsize{$\pm$ 0.031} \\
 & Median & 0.709 \scriptsize{$\pm$ 0.038} & 0.559 \scriptsize{$\pm$ 0.029} & 0.582 \scriptsize{$\pm$ 0.014} & 0.563 \scriptsize{$\pm$ 0.031} \\
 & WWAggr & \textbf{0.773} \scriptsize{$\pm$ 0.011} & \textbf{0.621} \scriptsize{$\pm$ 0.043} & \textbf{0.618} \scriptsize{$\pm$ 0.036} & \textbf{0.564} \scriptsize{$\pm$ 0.052} \\
\hline
\multirow{6}{*}{Road Accidents} & None (s.m.) & 0.336 \scriptsize{$\pm$ 0.024} & 0.319 \scriptsize{$\pm$} 0.032 & 0.359 \scriptsize{$\pm$} 0.017 & 0.361 \scriptsize{$\pm$} 0.020 \\ \cdashline{2-6}
 & Mean & \underline{0.354} \scriptsize{$\pm$ 0.007} & 0.317 \scriptsize{$\pm$ 0.023} & \underline{0.381} \scriptsize{$\pm$ 0.014} & \underline{0.379} \scriptsize{$\pm$ 0.009} \\
 & Min & 0.337 \scriptsize{$\pm$ 0.031} & 0.302 \scriptsize{$\pm$ 0.028} & 0.354 \scriptsize{$\pm$ 0.008} & 0.354 \scriptsize{$\pm$ 0.008} \\
 & Max & 0.353 \scriptsize{$\pm$ 0.031} & 0.317 \scriptsize{$\pm$ 0.020} & 0.364 \scriptsize{$\pm$ 0.031} & 0.364 \scriptsize{$\pm$ 0.031} \\
 & Median & 0.351 \scriptsize{$\pm$ 0.009} & \underline{0.337} \scriptsize{$\pm$ 0.028} & 0.378 \scriptsize{$\pm$ 0.011} & 0.378 \scriptsize{$\pm$ 0.011} \\
 & WWAggr & \textbf{0.383} \scriptsize{$\pm$ 0.024} & \textbf{0.407} \scriptsize{$\pm$ 0.007} & \textbf{0.391} \scriptsize{$\pm$ 0.022} & \textbf{0.384} \scriptsize{$\pm$ 0.039} \\
\hline
\end{tabular}
\end{table*}

\subsection{Base change point detectors}
\label{subsec:base_change_point_detectors}
We prove the versatility of the proposed pipeline by conducting experiments with various SOTA deep change point detectors as ensemble base learners.
As \emph{supervised} standalone detectors, we stick to the generic classification-based BCE~\cite{lattari2022deep, romanenkova2022indid} model and CPD-specific InDiD model from~\cite{romanenkova2022indid}.
For \emph{unsupervised} variants, we experiment with the TS-CP$^{2}$ model presented by~\cite{deldari2021time} and SN-TS2Vec developed by~\cite{bazarova2024norm}.
Please find brief descriptions of these models in Appendix~\ref{sec:impl_details} or look at the original papers for more details.

\subsection{Datasets}
\label{subsec:datasets}
Although our primary focus is on high-dimensional data CPD, we also evaluate our approaches on simpler sequences. 
For this, we selected the \emph{Yahoo!} dataset~\citep{laptev2015s5}, a widely recognized CPD benchmark. 
Its diverse collection of univariate series with changes in trends, seasonality, and noise regimes presents a sufficient challenge even for deep CP detectors~\cite{chang2018kernel, deldari2021time, bazarova2024norm}. 
As more complicated datasets, we use high-dimensional video data: \emph{Explosions} and \emph{Road Accidents}, presented in~\citep{romanenkova2022indid, sultani2018real}. 
Both of them consider detecting sophisticated changes of varying intensities, which are complicated by the significant noisiness of the change signal and a wide diversity of normal behavior patterns.
The main datasets' characteristics are summarized in Table~\ref{tab:datasets}, while the detailed description is postponed to Appendix~\ref{sec:dataset_description}.

\begin{table}[H]
\caption{Statistics of the datasets used in the experiments.}
\label{tab:datasets}
\begin{adjustbox}{width=0.99\linewidth}
\begin{tabular}{l|ccc}
\hline
\multirow{2}{*}{Dataset}  & \multirow{2}{*}{Yahoo} & \multirow{2}{*}{Explosions} & Road \\
 & & & Accidents \\ \hline
Single sample shape & 1 & $ (240, 320, 3)$ & $ (240, 320, 3)$ \\
Train / Test seq. length & 150 / 1000 & 16 / 16 & 16 / 16 \\
Train / Test size & 21420 / 420 & 310 / 315 & 666 / 349 \\
\% train / test seq. with CP & 20.1 / 89.8 & 50.0 / 4.8 & 50.0 / 14.0 \\
\hline
\end{tabular}
\end{adjustbox}
\end{table}


\subsection{Main results}
\label{subsec:main_results}
Following previous works~\cite{romanenkova2022indid, chang2018kernel, deldari2021time, bazarova2024norm}, CPD models are evaluated using the standard $F_1$-score, which is based on task-specific interpretation of the confusion matrix elements.
Note that here, we are comparing not models themselves but various prediction aggregation techniques within each ensemble.
As ``naive'' procedures, we consider point-wise ``mean'', ``min'', ``max'', and ``median'' aggregations~\cite{katser2021unsupervised}.

Table~\ref{tab:results_ens} presents the main results for the supervised and unsupervised CPD ensembles. 
The average performance ranks for all the considered aggregations are presented in Table~\ref{tab:ranks}.
Based on the results, we conclude that:
\begin{enumerate}
    \item Ensembles of deep change point detectors consistently outperform standalone models. Confirming prior results~\cite{qin2025pca, katser2021unsupervised}, even simple aggregation is sufficient for low-dimensional Yahoo. However, for high-dimensional video data, these naive procedures are suboptimal, while our WWAggr provides superior performance.
    \item Our WWAggr technique is model-agnostic, working effectively with ensembles of various supervised and unsupervised deep CPD models. 
    In almost all cases, it boosts the performance.
    \item For the most challenging datasets (Explosions and Road Accidents), the proposed approach significantly improves the detection quality by increasing the detection $F_1$-score by up to 20\%.
\end{enumerate}

Considering these findings, we study the effect of model calibration on the issue of alarm threshold selection.

\subsection{Model calibration and alarm threshold selection}
\label{subsec:results_calibration}

\begin{figure*}[ht]
\centering
\includegraphics[width=0.75\linewidth]{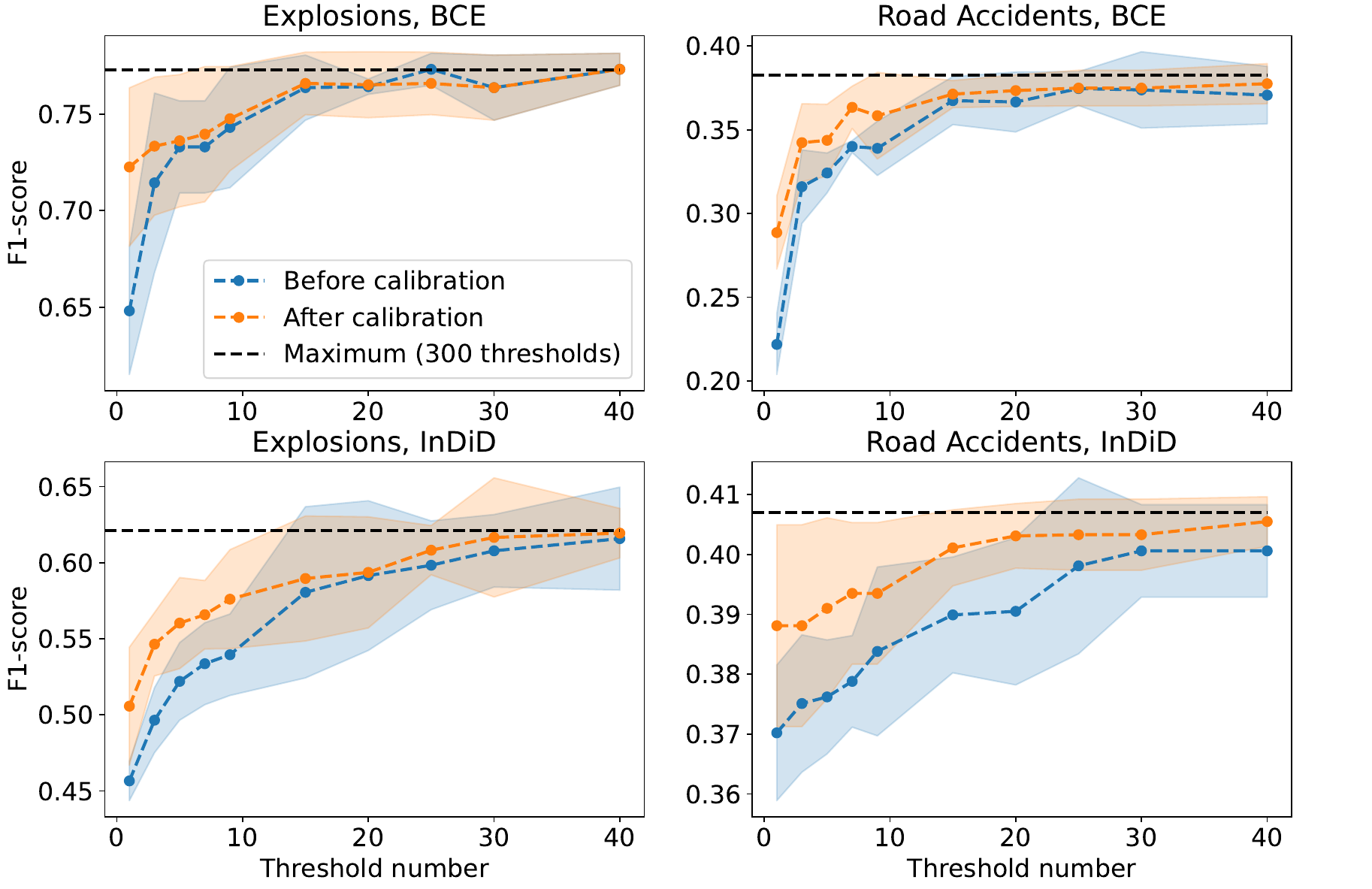}
\caption{
Dependence of the $F_1$-score on threshold selection for the WWAggr procedure before and after model calibration.
Dashed line indicates the best score obtained by searching through $300$ thresholds.
}
\label{fig:calibration_th_metrics}
\end{figure*}

As outlined in Section~\ref{subsec:ensemble_prediction_agregation}, the performance of seq2seq CPD models is often threshold-dependent.
This means one should carefully select the value of $h$ to obtain the best quality, which may be restrictive in practice.
During the experiments, we discovered that poor calibration of the ensemble's base learners results in higher sensitivity for the threshold selection. 

Figure~\ref{fig:calibration_th_metrics} compares maximum detection F1-scores obtained for various threshold numbers.
According to the results, 
we do not need to search through a large set of thresholds for WWAggr after the proper calibration.
Moreover, even 1 --- 5 values of $h$ are enough to obtain high-quality detection.
Such an effect allows the user to select the threshold in advance (e.g., $h = 0.5$), facilitating online inference and reducing the computational costs.

\subsection{Sensitivity studies}
\label{subsec:sensitivity_studies}
In this section, we provide our findings on how different components of the pipeline (ensembling strategies, calibration methods, and ensemble structure) affect the performance.
Due to space limitations, we present the results only for the ensembles of BCE models trained on the Explosions dataset.
However, this generalizes to the rest of the model-dataset pairs under study.

\subsubsection{Choice of ensembling method}
\label{subsubsec:sens_ensembling_method}
First, we explore how different ensembling techniques influence the overall quality of the proposed CPD method.
The models in the ensemble are supposed to be different to increase the resulting ensemble expressiveness, and this variety can be achieved in several ways~\citep{ganaie2022ensemble}. 
In particular, we compare three options:
\begin{enumerate}
    \item[(a)] \textit{Naive}: models in one ensemble differ only in weight initializations and random seeds for the SGD.
    \item[(b)] \textit{Bootstrap}: models in one ensemble are trained on different bootstrapped subsamples of the original train dataset.
    \item [(c)] \textit{SGLD}: models in one ensemble are trained with the noise-injection procedure proposed by~\cite{welling2011bayesian}.
\end{enumerate}
The results are presented in Table~\ref{tab:sens_ensembling_method_results}.
Based on these results, we conclude that the simplest (``naive'') approach provides enough model diversity and ensemble expressiveness for our task.
Bootstrapping different training subsamples for different base learners, as well as using the SGLD procedure, results in inferior performance for all the considered aggregation techniques. 
The reason for that is the data complexity: even SOTA standalone models are not able to converge to one global optimum, and additional randomization in the training process occurs to be redundant.
Therefore, we use simple ensembles of type (a) in our main experiments.

\begin{table}[ht]
\caption{
Detection $F_1$-scores for different ensembling methods, supervised BCE models trained on the Explosions dataset.
The results are averaged by three runs.
The best values are highlighted in \textbf{bold}.
}
\label{tab:sens_ensembling_method_results}
\centering
\begin{tabular}{l|ccc}
\hline
 \multirow{2}{*}{Aggregation} & \multicolumn{3}{c}{Ensembling method} \\ \cline{2-4}
 & Naive &  Bootstrap & SGLD \\
\hline
 Mean & \textbf{0.701} \scriptsize{$\pm$ 0.052} & 0.670 \scriptsize{$\pm$ 0.062} & 0.692 \scriptsize{$\pm$ 0.040} \\
 Min & \textbf{0.679} \scriptsize{$\pm$ 0.047} & 0.641 \scriptsize{$\pm$ 0.026} & 0.645 \scriptsize{$\pm$ 0.051} \\
 Max & \textbf{0.735} \scriptsize{$\pm$ 0.050} & 0.689 \scriptsize{$\pm$ 0.071} & 0.690 \scriptsize{$\pm$ 0.039} \\
 Median & \textbf{0.709} \scriptsize{$\pm$ 0.038} & 0.678 \scriptsize{$\pm$ 0.038} & 0.667 \scriptsize{$\pm$ 0.052} \\
 WWAggr & \textbf{0.773} \scriptsize{$\pm$ 0.011} & 0.728 \scriptsize{$\pm$ 0.034} & 0.741 \scriptsize{$\pm$ 0.043} \\
\hline
 &\multicolumn{3}{c}{$F_1$ for a standalone model is 0.695 \scriptsize{$\pm$ 0.058} }\\ \hline
\end{tabular}
\end{table}

\subsubsection{Mixed ensembles}
\label{subsubsec:sens_mixed_ensembles}
Additionally, we prove the versatility of the WWAggr procedure by experimenting with the mixed ensembles. 
In more detail, we construct two types of mixed ensembles: supervised (consisting of five BCE and five InDiD models) and unsupervised (with five TS-CP$^2$ and five SN-TS2Vec base learners). 
The results are presented in Table~\ref{tab:sens_mixed_ensembles_results}.
As expected, we observe that the performance quality of the mixed ensembles lies between the corresponding homogeneous ensembles (Section~\ref{subsec:main_results}).
However, it is more important to note that our WWAggr aggregation technique remains superior in this case as well, thereby proving its universality and robustness to the underlying ensemble structure.

\begin{table}[ht]
\caption{
Detection $F_1$-scores for the mixed ensembles of supervised and unsupervised base learners, trained on the Explosions dataset.
The results are averaged by three runs.
The best values are highlighted in \textbf{bold}.
}
\label{tab:sens_mixed_ensembles_results}
\centering
\begin{tabular}{l|c|c}
\hline
\multirow{2}{*}{Aggregation} & \multicolumn{2}{c}{Ensemble structure} \\ \cline{2-3}
 & BCE + InDiD & TS-CP$^2$ + SN-TS2Vec \\ \hline
Mean & 0.675 \scriptsize{$\pm$ 0.015} & 0.499 \scriptsize{$\pm$ 0.062} \\
Min & 0.657 \scriptsize{$\pm$ 0.057} & 0.538 \scriptsize{$\pm$ 0.002} \\
Max & 0.673 \scriptsize{$\pm$ 0.060} & 0.549 \scriptsize{$\pm$ 0.023} \\
Median & 0.676 \scriptsize{$\pm$ 0.029} & 0.568 \scriptsize{$\pm$ 0.033} \\
WWAggr & \textbf{0.684} \scriptsize{$\pm$ 0.032} & \textbf{0.587} \scriptsize{$\pm$ 0.024} \\
\hline
\end{tabular}
\end{table}

\subsubsection{Choice of calibration method}
\label{subsubsec:sens_calibration_method}
\begin{figure*}[ht]
\centering
\includegraphics[width=0.80\textwidth]{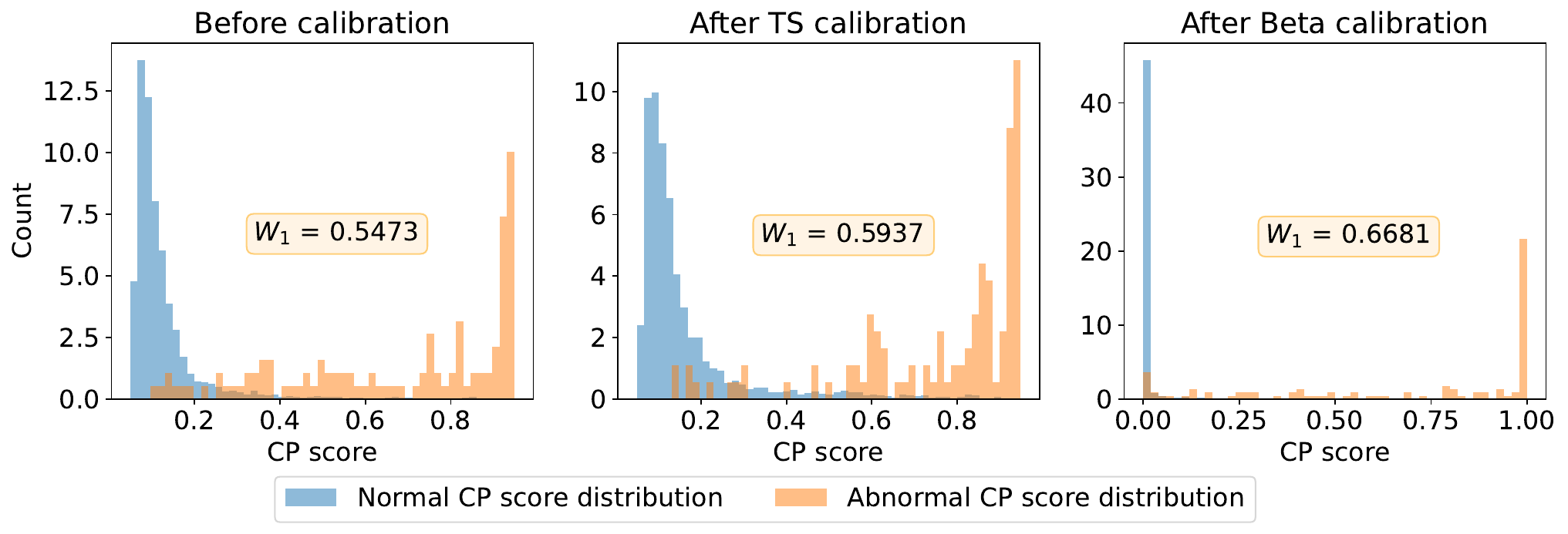}
\caption{
Histograms of the mean predicted ``normal'' and ``abnormal'' CP scores for ensembles of supervised BCE models trained on the Explosions dataset.
$W_{1}$ represents the 1-Wasserstein distance estimate between these two distributions.
}
\label{fig:calibration_effect}
\end{figure*}

In this study, we compare two popular post-hoc calibration techniques: the Temperature Scaling~\citep{guo2017calibration} and the beta calibration~\citep{kull2017beyond}.
Table~\ref{tab:mean_ece} presents values of Expected Calibration Error (ECE) averaged across ten models in the ensemble.
According to the results, Temperature Scaling, with only one tunable parameter, lacks the flexibility to significantly improve model calibration.
On the contrary, the Beta method allows to achieve almost perfect ECE values.
Figure~\ref{fig:calibration_effect} complements this study and reports the histograms of mean ``normal'' (before a CP) and ``abnormal'' (after a CP) change point scores predicted by ensembles of BCE models for the Explosions dataset.
It is evident that a proper model calibration enables better separation of these two CP score distributions, which results in a larger 1-Wasserstein distance between them.
Taking these findings into account, we settled on the beta method as the main option for the calibration step in the proposed pipeline.

\begin{table}[!ht]
\caption{
Expected Calibration Error (see~\citep{guo2017calibration}; should be minimized $\downarrow$) values for the BCE models trained on the Explosions dataset before and after calibration.
The best metric value is highlighted in \textbf{bold}.
}\label{tab:mean_ece}
\centering
\begin{tabular}{l|c}
\hline
Calibration method & Mean ECE $\downarrow$\\ \hline
None & 0.1185 \\
Temperature Scaling & 0.0931 \\
Beta & \textbf{0.0023} \\ \hline
\end{tabular}
\end{table}

\subsubsection{Choice of probabilistic distance}
\label{subsubsec:choice_of_probabilistic_distance}
\begin{figure}[ht]
\centering
\includegraphics[width=0.75\linewidth]{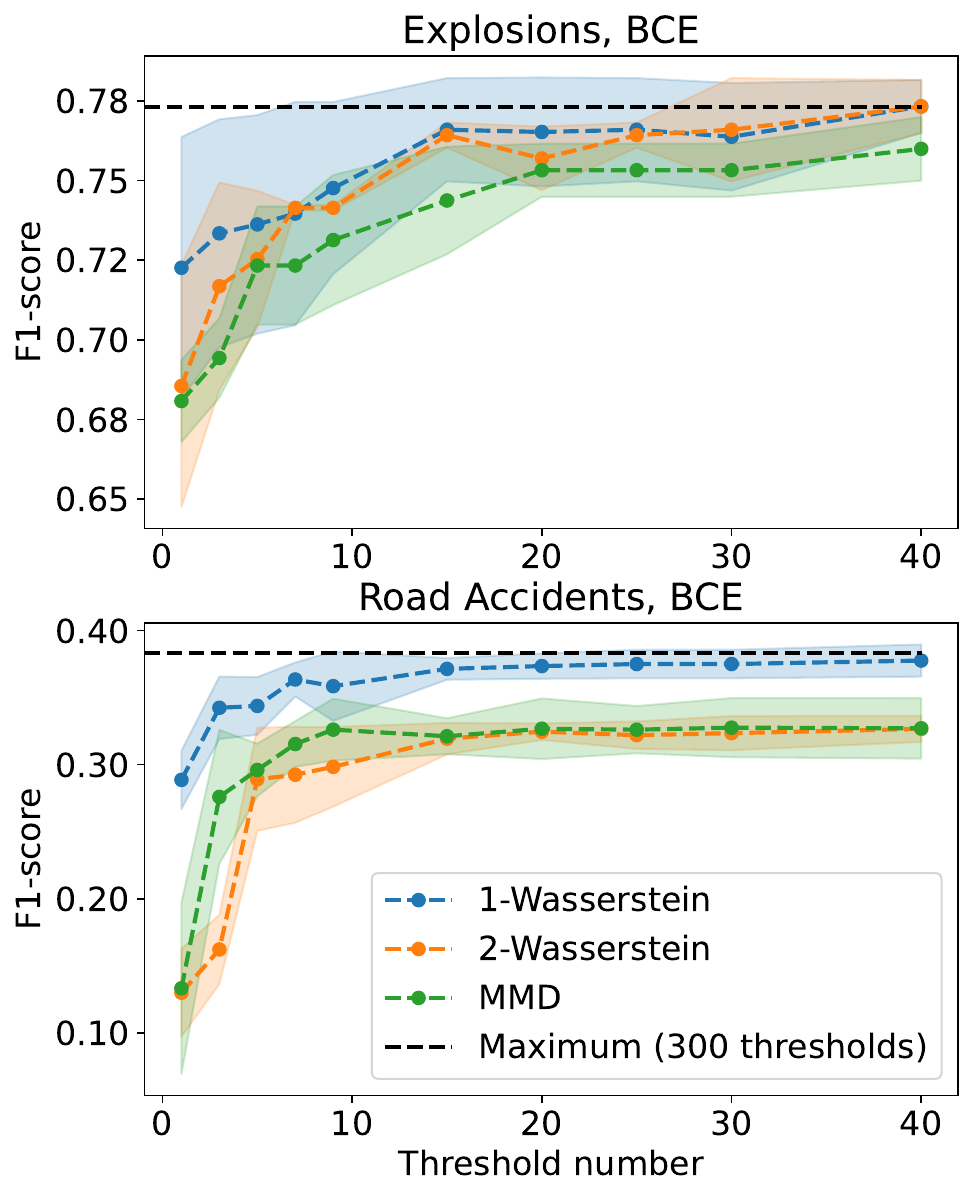}
\caption{
Dependence of the $F_1$-score on threshold selection for our aggregation procedure with different probabilistic distances.
Results for the experiments with calibrated BCE ensembles on the video datasets.
Dashed line indicates the best score obtained by searching through $300$ thresholds.
}
\label{fig:th_range_distances}
\end{figure}

In this experiment, we examine the sensitivity of the proposed aggregation technique from Algorithm \ref{alg:distance_procedure} to the choice of the probabilistic distance $d(\cdot)$.
Although all three considered options ($\hat{W}_1$, $\hat{W}_2$, MMD) produce close detection quality when the alarm threshold is carefully selected, the 1-Wasserstein distance behaves in a more stable way with 1 --- 5 standard predefined thresholds (see Figure~\ref{fig:th_range_distances}).
Thus, this is the most empirically appropriate distance variant for WWAggr.

\section{Limitations and future work}
\label{sec:limitations}
The main limitation of our approach is its computational complexity.
On the Explosions dataset, the average GPU inference time per sequence for BCE/InDiD increases from $40.5\pm 1.2$ ms for a single model to $400.0\pm 3.2$ ms for an ensemble of size $K = 10$ with ``naive'' aggregations.
Our WWAggr method with the same ensemble size took 
$417.1\pm 5.0$ ms, incurring a minor computational overhead ($\approx 5\%$). 
WWAggr has the potential to overcome this difference and be more efficient if the sliced~\citep{kolouri2019generalized} or projected~\citep{wang2021two} Wasserstein distances are used.

Additionally, this work targets high-dimensional data, where assumption-free deep CP detectors are the only reasonable solution~\cite{romanenkova2022indid, ryzhikov2023latent}. 
However, for lower dimensions, the increase in computational cost for ensembles of such deep detectors may outweigh its modest performance benefits. 
In this case, one could resort to Bayesian Neural Networks~\cite{jospin2022hands}, although their efficacy strongly depends on making proper prior choices~\cite{lakshminarayanan2017simple}.
Another alternative could be to ensemble classical statistical methods, such as CUSUM~\cite{shiryaev2017stochastic}. 
While we hypothesize our aggregation strategy would also be effective for these procedures, classical methods eliminate the calibration of CP scores and, consequently, any guarantee for easy alarm threshold choice.
Thus, we leave these ideas for future investigation.
\section{Conclusion}
\label{sec:conclusion}
We tackled the problem of change point detection in real-world high-dimensional data, where traditional methods often underperform due to complex data patterns. 
Recognizing the limitations of standalone deep neural networks, the paper proposes using ensembles of SOTA change point detectors, which are more suitable for real-world challenges.

In our experiments, ensembles of deep CPD detectors consistently outperform single models, even with simple aggregation methods like averaging. 
However, these ``naive'' procedures are sub-optimal for the change point detection setting.
Thus, we introduce WWAggr, the novel ensemble prediction aggregation procedure based on the Wasserstein distance.
This procedure significantly boosts detection quality by capturing the peculiarities of the CPD task, being model-agnostic and suitable for various supervised and unsupervised CPD base learners.
As a result, WWAggr advances an ensemble, improving the detection $F_1$-score on the video data by up to $20\%$ compared to basic aggregations.

To ensure robustness against alarm threshold selection, we apply model calibration, making the proposed approach more practical.
Consequently, in inference, almost any reasonable threshold value results in a decent performance quality.


\section*{Acknowledgements} 

The work was supported by the grant for research centers in the field of AI provided by the Ministry of Economic Development of the Russian Federation in accordance with the agreement 000000C313925P4F0002 and the agreement with Skoltech №139-10-2025-033.
{\small
\bibliographystyle{ieeetr}
\bibliography{bib}
}

\appendices
\section{Dataset descriptions}
\label{sec:dataset_description}
\paragraph{Yahoo}
The Yahoo! dataset~\cite{laptev2015s5} contains univariate time series with varying trends, seasonality, and noise regimes.
Originally proposed for benchmarking the Anomaly Detection methods, this dataset has been adapted to change point detection by, e.g.,~\cite{chang2018kernel, deldari2021time, bazarova2024norm}.
Following these works, we use $100$ long time series from the fourth benchmark, as this is the only portion of the dataset that includes CPs but not anomalies.
We further extract $420$ sequences of length $1000$ for the testing subset and use around $20000$ sequences of length $150$ for training.
For LSTM-based supervised base learners (BCE and InDiD), we cropped test sequences to a length of 150, keeping the proportion of ``normal'' and ``abnormal'' series the same.

\paragraph{Explosions and Road Accidents}
As more complicated datasets, we use high-dimensional semi-structured video data.
To construct these samples, real-world RGB videos were taken from the UCF-Crime anomaly detection dataset~\citep{sultani2018real}, and their CPD markup was provided by~\cite{romanenkova2022indid}.
In the \emph{Explosions} dataset, CPs correspond to any explosions and fire shot by a CCTV camera.
"Normal" videos do not include any emergencies.
These situations follow the CPD problem statement as the explosion consequences (fire, smoke, etc.) are seen for some time afterwards.
The clips from \emph{Road Accidents} follow a similar pattern but are more challenging.
In this case, CPs indicate car crashes on the streets.
As traffic accidents do not necessarily split the video into two regimes, we use this dataset to measure the range of applicability of CPD models.
In these two video datasets, every clip has a length of $16$ frames, and every frame is a tensor of size $240\times 320\times 3$.

\section{Implementation details}
\label{sec:impl_details}
We conducted all the experiments from the paper using Python 3 and the PyTorch framework.
The code is available online\footnote{\url{https://github.com/stalex2902/wwagr-ensemble-cpd}}.
All the models were trained on two NVIDIA GeForce GTX 1080 Ti GPUs.
For the implementation of base deep CPD models, which form various ensembles, we mostly followed the original works and inherited the proposed model architectures, which is describes below.
Additionally, we provide details on implementing the proposed sliding-window Wasserstein-based aggregation procedure.

\subsection{Details on supervised ensembles}
\label{subsec:supervised_cpd}
We utilize two sequence-to-sequence supervised approaches. Both models employ recurrent LSTM-based~\citep{hochreiter1997long} architectures but are trained with different loss functions. 
The first one, denoted as BCE, is a generic neural network that considers CPD as a classification task and directly optimizes the standard binary cross-entropy loss~\cite{lattari2022deep,  romanenkova2022indid, hushchyn2020online}. 
In contrast, the second one, InDiD model, leverages the principles of the CPD task and optimizes the detection delay and number of false alarms~\cite{romanenkova2022indid}. 
Both approaches are recognized for their high quality across various real-world CPD tasks, including the detection of changes in video samples.
We refer the reader to the original paper for more details.

These BCE and InDiD models consist of LSTM~\citep{hochreiter1997long} blocks followed by dense layers.
To stabilize this model, we added Layer Norm~\cite{ba2016layernormalization} to the LSTM cells, which resulted in more robust training and better final metrics compared to the original paper~\cite{romanenkova2022indid}.
The rest of the hyper-parameters and the training pipeline were taken from the repository\footnote{\url{https://github.com/romanenkova95/InDiD}}.

\subsection{Details on unsupervised ensembles}
\label{subsec:unsupervised_cpd}
Additionally, we study ensembles of unsupervised models TS-CP$^{2}$~\citep{deldari2021time} and SN-TS2Vec~\citep{bazarova2024norm}.
These two models are inspired by contrastive self-supervised learning techniques.

The general idea of TS-CP$^{2}$~\citep{deldari2021time} is to train a CNN using pairs of "similar" (positive) and "dissimilar" (negative) data windows with the contrastive loss function, adapting the CPC approach~\cite{oord2018representation}. 
As a result, the representations of the positive pairs are brought closer, and the embeddings of the negative pairs are further apart. 
Thus, change points can be detected by computing the cosine distance between the representations of consecutive windows of a time series. TS-CP$^{2}$ model has a one-dimensional convolutional architecture with the hyper-parameters mentioned in the work of~\cite{deldari2021time}.
We re-implement the model in PyTorch based on the original code from the repository\footnote{\url{https://github.com/cruiseresearchgroup/TSCP2}}.

SN-TS2Vec~\cite{bazarova2024norm} is the improvement of the prominent TS2Vec model~\cite{yue2022ts2vec}, which is also based on convolutional networks.  
It utilizes the spectral normalization technique, enabling the CPD-friendly embedding properties.
The final CPD predictions are obtained in the same way as for the TS-CP$^{2}$ model. We strictly adhere to the original implementation\footnote{\url{https://gitlab.com/abazarova/ssl_cpd}}, including their choice of architecture and training strategies.

Note that both types of unsupervised base learners (TS-CP$^{2}$ and SN-TS2Vec) are based on representation learning.
Consequently, they detect change points by computing the cosine distances between subseries representations, and these distances, bounded to $[-1, 1]$, serve as CP scores $p_{t}$.
On the other hand, supervised baselines operate with CP probabilities $p_{t}\in [0, 1]$.
In order to align these two approaches, we suggest post-processing TS-CP$^{2}$ and SN-TS2Vec outputs with the following transformation: $\tilde{p}_{t} = \left(1 - p_{t}\right)\cdot\mathbb{I}\left(p_{t} \geq 0\right)$, where $\mathbb{I}(\cdot)$ represents the indicator function.
As a result, the final CP scores lay in the range from 0 to 1, allowing us to apply WWAggr in all the cases.

\subsection{Video data preprocessing}
\label{subsec:video_cpd}
Following~\cite{romanenkova2022indid}, we preprocess the video clips from Explosions and Road Accidents with a pre-trained feature extractor X3D~\citep{feichtenhofer2019slowfast}.
This CNN-based model takes an input RGB video as a tensor of shape $320\times 240\times 3\times T$ and outputs its feature tensor of shape $12288\times T$. 
Therefore, it reduces data dimensionality and extracts informative features for each frame that reflect the distribution changes in the original video.
Extractor parameters were frozen while training the models.

\subsection{Hyper-parameter selection}
\label{subsec:hyperparameter_selection}
In all the experiments, we used ensembles of size $K = 10$.
To find the optimal values for the window size parameter $\omega$ of the proposed aggregation technique, we evaluate several options on the validation set.
In more detail, we search through $\omega\in [1, 2, 3]$ for the Explosions and Road Accidents datasets, and $\omega \in [3, 5, 10]$ for the Yahoo dataset.
For each dataset-model pair, the best result is reported.




\end{document}